\documentclass[letterpaper, 10 pt, conference]{ieeeconf}  % Comment this line out if you need a4paper

\IEEEoverridecommandlockouts                              % This command is only needed if 
\usepackage{epsfig}
\usepackage{makecell}
\usepackage{subcaption}
\usepackage{xcolor}
\usepackage{ragged2e}
\usepackage{float} 

\usepackage{graphicx}
\usepackage{amsmath}
\usepackage{adjustbox}
\usepackage{amssymb}
\usepackage{booktabs}
\usepackage{tabularx}
\usepackage{array}
\usepackage{makecell}
\usepackage{bm}

\title{\LARGE \bf
Human Mobility Modeling with Household Coordination Activities under Limited Information via Retrieval-Augmented LLMs
}

\author{Yifan Liu$^{1}$, Xishun Liao$^{1,*}$, Haoxuan Ma$^{1}$, Brian Yueshuai He$^{2}$, Chris Stanford$^{3}$, and Jiaqi Ma$^{1}$% <-this % stops a space
\thanks{$^{1}$UCLA Mobility Lab, Department of Civil and Environmental Engineering, University of California, Los Angeles, Los Angeles, USA.}
\thanks{$^{2}$Civil and Environmental Engineering Department, University of Louisville, Louisville, Kentucky, USA.}
\thanks{$^{3}$Novateur Research Solutions, Ashburn, VA, USA.}
\thanks{*Corresponding author: xishunliao@ucla.edu}
}

\begin{document}

\maketitle
\thispagestyle{empty}
\pagestyle{empty}

%%%%%%%%%%%%%%%%%%%%%%%%%%%%%%%%%%%%%%%%%%%%%%%%%%%%%%%%%%%%%%%%%%%%%%%%%%%%%%%%
\begin{abstract}
  Understanding human mobility patterns has long been a challenging task in transportation modeling. Due to the difficulties in obtaining high-quality training datasets across diverse locations, conventional activity-based models and learning-based human mobility modeling algorithms are particularly limited by the availability and quality of datasets. Current approaches primarily focus on spatial-temporal patterns while neglecting semantic relationships such as logical connections or dependencies between activities and household coordination activities like joint shopping trips or family meal times, both crucial for realistic mobility modeling. We propose a retrieval-augmented large language model (LLM) framework that generates activity chains with household coordination using only public accessible statistical and socio-demographic information, reducing the need for sophisticated mobility data. The retrieval-augmentation mechanism enables household coordination and maintains statistical consistency across generated patterns, addressing a key gap in existing methods. Our validation with NHTS and SCAG-ABM datasets demonstrates effective mobility synthesis and strong adaptability for regions with limited mobility data availability.

\end{abstract}

%%%%%%%%%%%%%%%%%%%%%%%%%%%%%%%%%%%%%%%%%%%%%%%%%%%%%%%%%%%%%%%%%%%%%%%%%%%%%%%%
\section{INTRODUCTION}

Understanding and accurately generating human mobility patterns remains a fundamental challenge in transportation research with implications for urban planning, public health, and even  retail strategies~\cite{barbosa2018human, li2021impact, vanhaverbeke2011agent}. Accurate mobility modeling can enhance transportation efficiency and urban design, ultimately improving civilian quality of life.

Traditional activity-based models (ABMs) have advanced our understanding of human mobility and travel demand by simulating daily activities based on socio-economic characteristics. Since their emergence in 1999, government agencies like Southern California Association of Governments (SCAG) have widely adopted these models for various applications including traffic analysis, urban planning, commercial strategy development, and tax policy evaluation~\cite{rasouli2014activity, bhat1999activity, goulias2011simulator, mcfadden1974measurement, heffer2021impact}. While ABMs effectively incorporate complex behavioral dynamics, they demand extensive local data and rely on numerous assumptions about human activity patterns and economic behaviors. Concurrently, data-driven approaches using neural networks have emerged to capture mobility patterns through large datasets from mobile phones and GPS~\cite{sila2016analysis, huang2018modeling, tang2015uncovering, liao2024deep}. 

However, these methods face several limitations, they require detailed individual travel diary data which raises privacy concerns, struggle to adapt to rapid urban changes, and rely on simplified behavioral assumptions that may not capture the flexibility of human decision-making in response to socio-economic changes~\cite{chakraborty2017interpretability, pellungrini2017fast}.

Recent advances in computational power have enabled Large Language Models (LLMs) to create new opportunities for human mobility modeling~\cite{ma2024mobility, liu2024semantic}. Models like GPT-3.5~\cite{achiam2023gpt} excel at generating human-like text across domains~\cite{NEURIPS2020_1457c0d6} and understanding complex sequences with strong interdependencies~\cite{NIPS2017_3f5ee243}. Trained on diverse textual data, LLMs can incorporate a wide range of human experiences and behaviors, potentially leading to more nuanced and varied human mobility modeling compared with conventional methods, as daily routines often involve intricate chains of activities with subtle interrelations. 

% Building upon the concept of ``activity chain", which represents the daily activity sequence of individual agents~\cite{liao2024deep}, our framework uses LLMs to generate daily activity sequences based on socio-demographic information of each member in a household and public activity statistics. We implement a retrieval-augmentation mechanism that ensures statistical consistency and enables realistic household coordination. This approach produces mobility patterns that align with agents' socio-demographics while maintaining real-world logic. %Our method achieves the lowest Jensen-Shannon Divergence (JSD) of $0.011$ when validated against NHTS data and SCAG-ABM results.

% By leveraging retrieval-augmented LLMs, we reduce dependence on extensive historical data and behavioral assumptions. This enables modeling complex mobility patterns across diverse geographic locations with limited data. Our approach generates highly consistent data with minimal hallucination, ensuring reliable outputs for practical applications. Through micro-simulation of daily activities, our approach could potentially enhance traffic demand modeling and support multi-scale transportation simulations for advanced urban planning~\cite{haystac2024}. 

Building on this potential, we introduce a novel application of LLMs for mobility data synthesis. Based on the concept of activity chains~\cite{liao2024deep}, which reflects daily sequences of individual activities, our framework employs retrieval-augmented LLMs to generate realistic, demographically consistent mobility patterns considering the household coordination. Given by socio-demographic attributes and public available statistics, the model ensures statistical consistency and coordinated behavior without relying on extensive historical data or detailed behavioral assumptions. This approach enables scalable mobility data synthesis across diverse regions with limited data availability, supporting micro-simulation and integrated transportation modeling for urban planning. Our study makes several key contributions to the field of human pattern modeling compared with existing literature:

\begin{figure*}[t]
  \centering
  \includegraphics[width=0.90\linewidth]{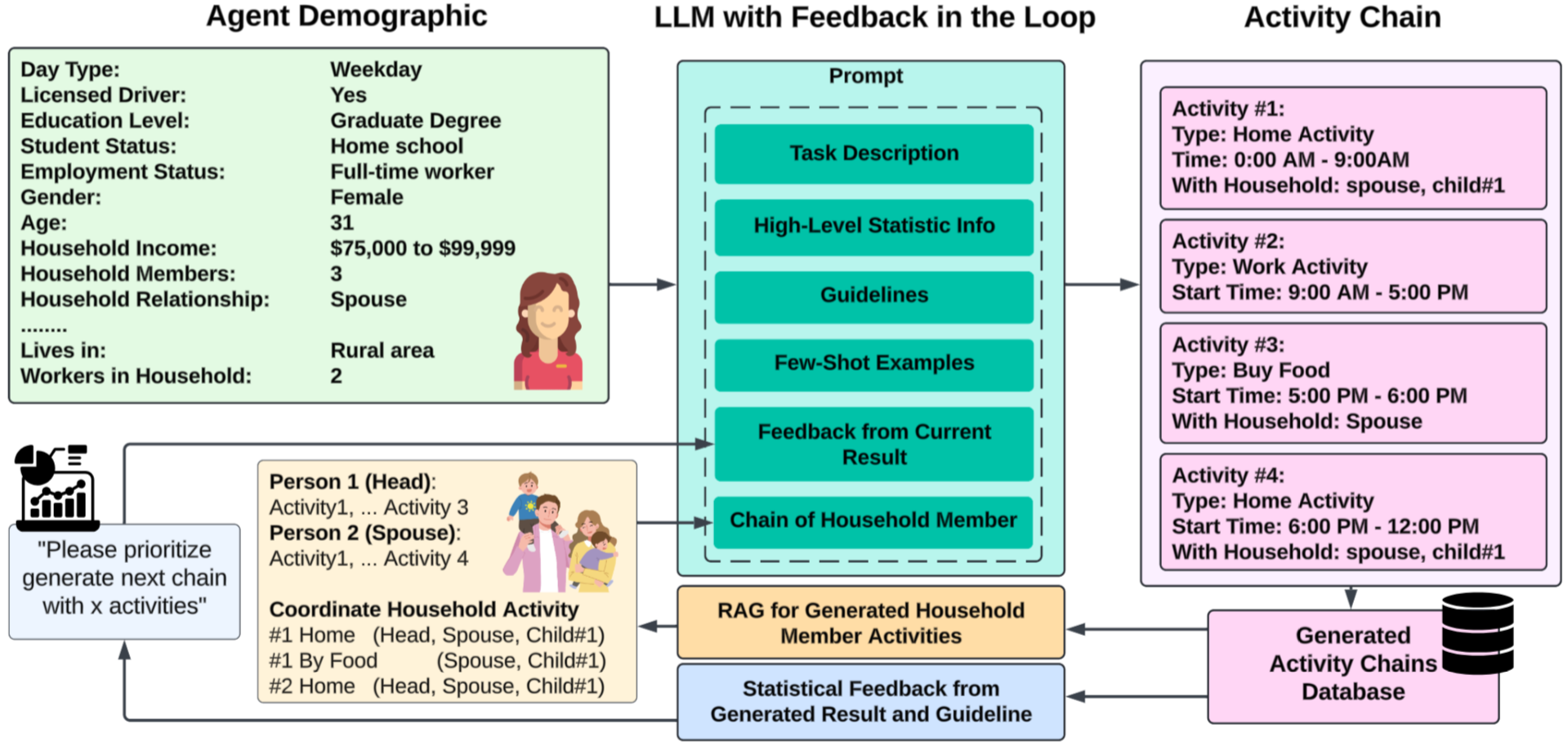}
  \caption{Proposed retrieval-augmented LLM framework with feedback loop for activity chain generation.}
  \label{fig:flow_diagram}
\end{figure*}

\begin{itemize}
    \item We propose a novel LLM-based approach for activity chain generation requiring only basic socio-demographic information and statistical data, eliminating dependency on extensive mobility datasets and addressing privacy concerns.
    
    \item We develop the retrieval-augmented LLM framework with an integrated feedback loop mechanism that ensures statistical consistency in activity durations, frequencies, and temporal patterns across generated patterns, producing highly reliable and realistic results suitable for comprehensive large-scale mobility modeling applications. To he best of our knowledge, the proposed framework is the first applied retrieval-augmented LLM with feedback in human mobility data generation tasks.
    
    \item We pioneer the application of LLMs for household-coordinated activity generation, enabling realistic modeling of complex interdependent activities among household members, thereby addressing the gap in existing mobility modeling.
\end{itemize}

\section{LITERATURE REVIEW}

\subsection{Human Mobility Modeling}
Human mobility modeling has evolved significantly since the 1940s when the ``Law of Intervening Opportunities" first connected travel patterns to socio-economic factors~\cite{stouffer1940intervening, zipf1941national}. Modern GPS and electronic tracking technologies have enabled sophisticated data collection and generative modeling approaches, with activity chain generation becoming a key focus area. ABMs represent a major advancement in transportation planning by simulating individuals' daily activities~\cite{rasouli2014activity}, using socio-demographic data, land use information, and transportation networks to construct detailed activity chains~\cite{bhat2012household}. While SCAG implemented SimAGENT~\cite{bhat2013household} to analyze regional travel behaviors, ABMs require extensive data collection and rely heavily on assumptions about travel patterns, limiting their transferability.

Learning-based methods using deep learning \cite{kim2018method}, Graph Convolutional Networks \cite{kong2022exploring}, and transformers \cite{liao2024deep} offer alternatives when trained on mobility data from mobile devices \cite{hoteit2014estimating, yin2011diversified}, but remain dependent on high-quality data that is often expensive and restricted. Both traditional and learning-based approaches face limitations from their reliance on comprehensive datasets or numerous behavioral assumptions \cite{liao2024deep}, highlighting the need for training-free approaches like LLMs that can synthesize mobility patterns using more accessible data sources.

\subsection{Household Activity Modeling}
Early approaches treated household members as independent units, overlooking their natural interdependencies~\cite{srinivasan2005modeling}. Modern household activity modeling now incorporates intra-household interactions, jointly modeling in-home and out-of-home activities to capture trade-offs and interdependencies between members \cite{rezvany2023simulating}. Platforms like VirtualHome model complex household activities through atomic action sequences \cite{puig2018virtualhome}, while SMACH offers multi-agent simulations to study energy consumption patterns and behavioral impacts \cite{albouys2019smach}, collectively improving the realism of household activity models.

Despite advances, current models fail to adequately capture household schedule interdependencies, relying on rigid rules or extensive data that limits cross-cultural generalization. Computational demands also restrict scalability for large-scale implementations.

\subsection{Large Language Models}
LLMs trained on trillions of tokens have emerged as powerful tools with transformer architectures that excel across domains from personal assistance to vehicle navigation~\cite{li2024personal, mao2023gpt}. Their flexibility enables rapid adaptation to new scenarios with minimal input~\cite{achiam2023gpt}.

Retrieval-augmented generation (RAG) enhances LLMs by allowing access to external databases, grounding responses in reliable information~\cite{gao2023retrieval} and addressing hallucination issues, though challenges in trustworthiness remain~\cite{ni2025towards}. Complementary approaches use automated feedback loops to iteratively refine responses, reducing hallucinations across various tasks~\cite{yu2023refeed, peng2023check}.

Our framework leverages these technologies to generate realistic activity chains with minimal data requirements. By combining LLMs with specialized retrieval-augmentation, we ensure statistical consistency and enable household activity coordination without requiring extensive location-specific data, while capturing household activities' interdependencies.

\section{Methodology}
\subsection{Overview}
The problem addressed in this study is defined as generating the daily activity chain for individual agents based on their socio-demographic information throughout a day. For each agent $i$ with his or her socio-demographic information collection $D_i = \{{d_i}^1, {d_i}^2, \dots, {d_i}^n\}$, we aim to generate a daily activity chain $C_i$ for that agent where each activity in the chain is defined by its type $A$, start time $T_s$, end time $T_e$, and household members participating $H$. The output activity chain $C_i$ for agent $i$ can be represented as 

$C_i = [{A_i}^1, {T_{s,i}}^1, {T_{e,i}}^1, {H_i}^1], \dots, [{A_i}^n, {T_{s,i}}^n, {T_{e,i}}^n, {H_i}^n]$,

The architecture of our proposed framework is illustrated in Fig.~\ref{fig:flow_diagram}. The model's inputs consist of 9 representative socio-demographic attributes for each agent. This socio-demographic information feeds into the feedback loop LLM mechanism, which processes this data through several key modules: task description, high-level statistical information, generation guidelines, few-shot examples, and the critical feedback mechanism. The right section of Fig.~\ref{fig:flow_diagram} shows the output activity chains with their temporal structure and household coordination elements. Once generated, activity chains are stored in a database. RAG retrieved household-related info and statistical feedback from these results will then be used in the feedback loop to continuously refine subsequent activity chain generation tasks, ensuring consistency across household members and alignment with empirical distributions. 

% This integrated approach equips our framework with the necessary context to generate realistic and representative daily activity chains, leveraging LLMs' robust inferential capabilities while ensuring statistical consistency through the retrieval-augmented feedback loop. 

\begin{table}[h]
  \centering
  \caption{Activity types aggregated in the NHTS 2017 dataset for the Los Angeles area.}
  \begin{tabular}{|c|c|c|c|c|c|}
      \hline
      1 & Home & 2 & Work  & 3 & School \\ \hline
      4 & Caregiving & 5 & Buy goods & 6 & Buy services \\ \hline
      7 & Buy meals & 8 & General errands & 9 & Recreational \\ \hline
      10 & Exercise  & 11 & Visit friends & 12 & Health care \\ \hline
      13 & Religious & 14 & Something else & 15 & Drop off/Pick up\\ \hline
  \end{tabular}
  \label{table:activity_type_table}
\end{table}

Table~\ref{table:activity_type_table} presents the 15 activity types used in our framework, aggregated from the filtered NHTS 2017 dataset in Los Angeles area~\cite{fish2015transportation}. These categories encompass the full spectrum of daily human activities, from essential functions like home, work, and school to discretionary activities such as recreation, exercise, and social visits. This classification system provides a comprehensive foundation for generating realistic daily activity patterns.

\subsection{Socio-Demographic Information}
Agent socio-demographic information serves as the input to the model, ensuring the generated activity chains accurately reflect each agent's distinct characteristics. As illustrated in Fig.~\ref{fig:flow_diagram}, 9 socio-demographic attributes including gender, age, education level, student status, employment status, household relationships, income levels, driver license status, and geographic location info are represented using natural language descriptions. This conversion from structured socio-demographic data into natural language expressions facilitates deeper contextual comprehension by the LLMs, enhancing the realism and relevance of the generated daily activity patterns.

\begin{figure}
  \centering
  \includegraphics[width=0.95\columnwidth]{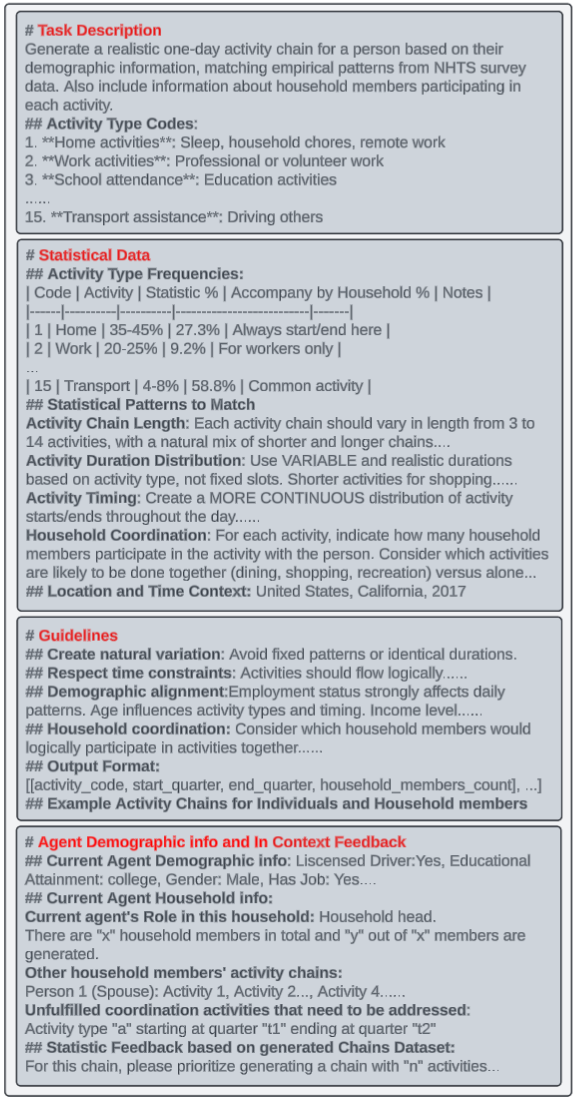}
  \caption{Example of input system prompt for LLMs.}
  \label{fig:llm_prompt}
\end{figure}

\subsection{Prompt Architecture}
We provide the LLMs with a structured system prompt that guides the generation of activity chains. The components of the system prompt are designed to provide comprehensive context and clear instructions, ensuring the generated outputs are both logically reasonable and aligned with the socio-demographic data. The structured input consists of the following elements:

The Large Language Model is guided through a structured prompt system, as detailed in Fig.~\ref{fig:llm_prompt}. This structured prompt integrates several critical components:

\begin{itemize}
  \item \textbf{Task Description:} Defines the goal explicitly, instructing the LLM to generate realistic daily activity chains for agents based on socio-demographic inputs and empirical patterns from datasets like the NHTS survey.

  \item \textbf{High-Level Statistical Information:} Provides statistical context, such as activity type frequencies, typical activity durations, household coordination probabilities, and spatial-temporal patterns, that the LLM uses to anchor generated activities to realistic empirical distributions.

  \item \textbf{Generation Guidelines:} Specifies constraints and standards for output realism, including the logical ordering of activities, duration distributions, and household coordination requirements. This ensures feasible and contextually coherent daily schedules.

  \item \textbf{Few-Shot Examples:} Demonstrates desired output formats through representative examples, helping the LLM internalize typical patterns of activity sequences, durations, and transitions. These examples enable the model to generate outputs adhering closely to observed human mobility patterns.

  \item \textbf{Retrieval-Augmented Generation Feedback:} Dynamically integrates statistical feedback and previously generated household member activities into the LLM's generation loop. This retrieval-augmented approach maintains consistency across household activity chains, supports realistic household coordination (such as shared home activities, joint travel, and synchronized activity timings), and aligns generated chains with observed statistical distributions.
\end{itemize}

Fig.~\ref{fig:llm_prompt} provides an example of the system prompt. These components collectively ensure that the LLM has a clear understanding of the task requirements and the contextual background needed to generate accurate and representative activity chains.

\subsection{Retrieval-Augmentation Mechanism}

Our framework incorporates a retrieval-augmentation mechanism to address limitations in sequential, agent-specific activity chain generation. LLMs typically struggle to maintain awareness of global statistical distributions when generating chains individually based on socio-demographic information. As shown in Fig.~\ref{fig:llm_prompt}, we overcome this by continuously monitoring and storing generated results, particularly tracking statistical attributes like activity chain lengths. This stored data provides real-time feedback to the LLM, guiding subsequent generations to maintain consistency with empirically observed distributions. The system continuously adapts prompts by analyzing previously generated data, guiding the LLM to produce activity chains that align with desired statistical distributions. Our experimental analysis shows that providing statistical feedback exclusively on chain length distribution successfully enhances both activity type and temporal distributions by utilizing the LLM's natural inference capabilities. This streamlined approach that focuses only on constraining chain length statistics minimizes potential bias while enabling the model to create naturally coherent patterns throughout various aspects of human mobility behavior.

For household coordination, our system implements a retrieval mechanism that incorporates previously generated activity patterns of other household members when generating chains for new members. The system retrieves existing household mobility patterns and pre-established coordination activities such as family meals, shared transportation, and joint recreational activities. This comprehensive household context enables the LLM to generate temporally aligned joint activities while maintaining schedule coherence, reflecting realistic family dynamics and interdependencies.

The mechanism proactively identifies and resolves scenarios where household coordination activities remain incomplete or unsynchronized, ensuring all activities are properly coordinated and temporally coherent. 

By integrating retrieval-augmentation with iterative feedback, our framework produces robust activity chains that reduce unrealistic mobility patterns brought by LLM hallucinations while ensuring statistically consistent and contextually appropriate household coordination.

\section{EXPERIMENT}

\subsection{Dataset}
\subsubsection{National Household Travel Survey Dataset}
The 2017 National Household Travel Survey (NHTS) dataset~\cite{NHTS2009} serves as our primary reference, providing comprehensive travel behavior data across the United States with detailed socio-demographic information and daily activity patterns. The survey collected data from approximately 129,000 households, encompassing about 264,000 persons, of which we utilized 180,000 high-quality person records after filtering. We aggregate activities into 15 types as shown in Table \ref{table:activity_type_table}. The NHTS dataset also captures household coordinated activities, allowing us to analyze and model the interdependencies of travel behaviors and activity patterns among household members.

\subsubsection{Activity-Based Model Dataset from Southern California Association of Governments}
We also utilize synthetic results from the Southern California Association of Governments (SCAG) Activity-Based Model~\cite{he2022connected, jiang2022connected}, which models travel demand for approximately 26 million people. This dataset provides weekday activity chains which we convert to match the activity type categorization in Table \ref{table:activity_type_table} for consistency and fair experimental comparison with our NHTS aggregation.

% \begin{figure*}[t]
%   \centering
%   \includegraphics[width=0.98\linewidth]{figures/overview.png}
%   \caption{Evaluation matrix on SCAG and NHTS dataset}
%   \label{fig:evaluation_overview}
% \end{figure*}

\subsection{Experiment and Result}

We evaluated our approach using the NHTS dataset in the California area by randomly sampling 500 agents and generating their daily activity chains based on their socio-demographic data. These chains were then validated against the comprehensive daily activity records from the NHTS dataset and the SCAG dataset. In our experiments, we utilized three large language models: OpenAI's GPT-4o mini, Meta's Llama3.1-70b, and DeepSeek v3, with the temperature setting of $1.0$ for each model, comparing their performance in accurately simulating daily human activities. The token size for input is around 700 and output around 4 on average for each activity chain generation. The generation speed for each instance was approximately 0.5 seconds for the GPT-4o mini and DeepSeek v3 API calls, and 1 second for the Llama3.1-70b model with 4-bit quantization running on an L40S GPU with 48GB memory.

The evaluation metrics will compare the distributions of activity type, start time, end time, duration, and the number of daily activities. We employed JSD~\cite{menendez1997jensen} to quantify the differences between the generated activity chains and the reference activity chains from the NHTS and the SCAG dataset. 

\begin{figure}[h]
  \centering
  \includegraphics[width=0.98\linewidth]{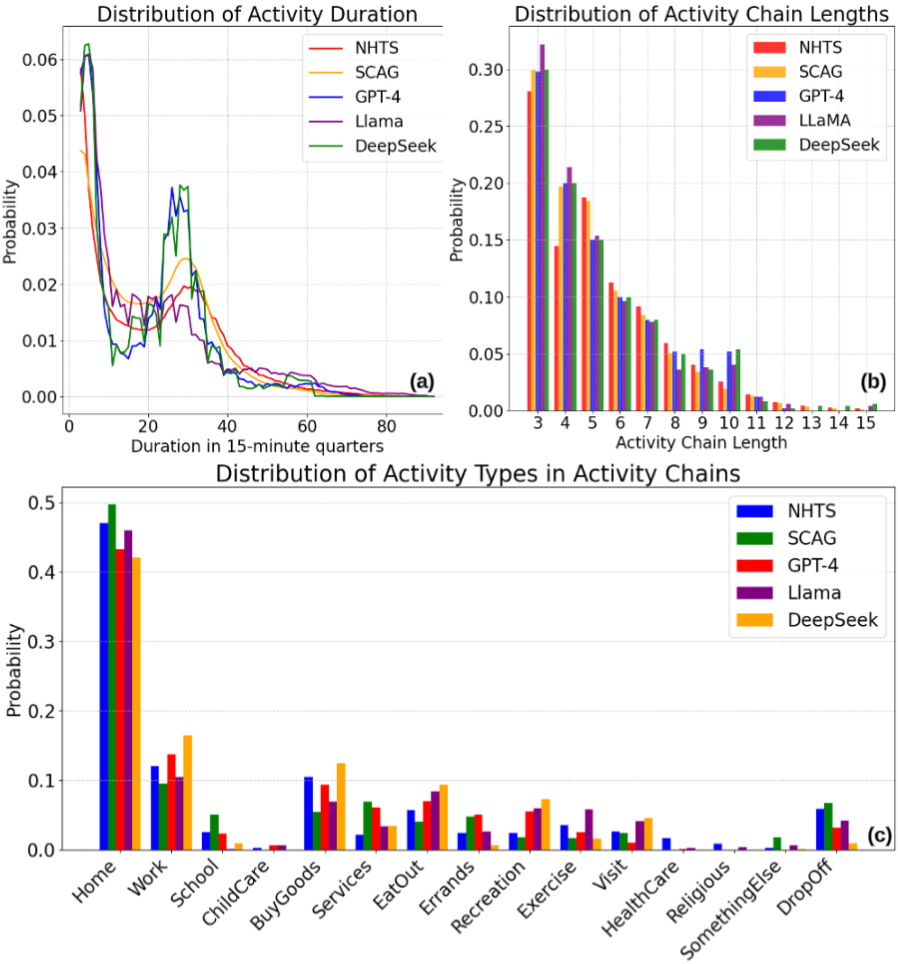}
  \caption{Evaluation matrix on SCAG and NHTS dataset}
  \label{fig:evaluation_overview_2}
\end{figure}

\begin{table}[h]
  \centering
  \renewcommand{\arraystretch}{1.2}
  \caption{JSD values comparing different models with reference datasets. Lower values indicate better alignment with reference distributions. The SCAG-NHTS comparison (bottom row) serves as a baseline, showing inherent differences between the two reference datasets.}
  \begin{tabular}{|>{\centering\arraybackslash}p{2.0cm}|>{\centering\arraybackslash}p{0.65cm}|>{\centering\arraybackslash}p{0.7cm}|>{\centering\arraybackslash}p{0.7cm}|>{\centering\arraybackslash}p{1.0cm}|>{\centering\arraybackslash}p{0.8cm}|}
  \hline
  Comparison & Type &  Time start & Time end & Duration & Length \\ \hline
  NHTS-SCAG & 0.031 & 0.007 & 0.006 & 0.006 & 0.003 \\ \Xhline{5\arrayrulewidth}
  NHTS-GPT & \textbf{0.023} & 0.015 & \textbf{0.011} & 0.022 & 0.011 \\ \hline
  NHTS-Llama & 0.024 & 0.090 & 0.081 & 0.023 & 0.011 \\ \hline
  NHTS-DeepSeek & 0.040 & \textbf{0.013} & 0.013 & 0.026 & \textbf{0.009} \\ \Xhline{3\arrayrulewidth}
  SCAG-GPT & \textbf{0.028} & \textbf{0.009} & \textbf{0.009} & \textbf{0.020} & 0.009 \\ \hline
  SCAG-Llama & 0.045 & 0.061 & 0.056 & 0.027 & \textbf{0.006} \\ \hline
  SCAG-DeepSeek & 0.065 & 0.013 & 0.016 & 0.020 & 0.007 \\ \hline
  \end{tabular}
  \label{table:jsd_comparison}
\end{table}

We analyzed the JSD values between our approach and the reference datasets, as detailed in Table~\ref{table:jsd_comparison}. A JSD value closer to $0$ indicates a more accurate approximation with the reference dataset's distribution. Overall, our LLM-based approach successfully captures the trends in human mobility patterns across all evaluated dimensions. Among the models we tested, GPT-4o mini demonstrates particularly strong performance, especially in activity type and end time modeling across both datasets. The JSD values between NHTS and SCAG datasets (ranging from 0.003 to 0.031) reveal inherent differences between survey-based data (NHTS) and synthetic data (SCAG-ABM), which provides important context for interpreting our results against each reference dataset.

\subsection{Activity Type}
Fig.~\ref{fig:evaluation_overview_2}$c$ shows our approach accurately represents activity type distributions from reference datasets, especially for common activities like home and work. Our models capture both frequent and rare activity types with varying accuracy. GPT-4o mini aligns particularly well with reference distributions, accurately representing less common activities such as childcare, healthcare, and errands.

\subsection{Activity Duration}
Fig.~\ref{fig:evaluation_overview_2}$a$ demonstrates our approach accurately models activity duration distributions from both reference datasets. The models capture the prevalence of shorter activities in NHTS and moderate-length activities in SCAG. All implementations perform well in this dimension, with GPT-4o mini showing the strongest alignment with both reference profiles, particularly for activities of various durations.

\subsection{Activity Chain Length}
Fig.~\ref{fig:evaluation_overview_2}$b$ shows our approach captures the preference for shorter chains (3-6 activities) present in both reference datasets. All models generate realistic activity chain lengths, though they struggle with complex chains exceeding eight activities. This limitation presents an opportunity for future improvement in modeling extended daily routines, while confirming our approach successfully captures common activity chain patterns.

\subsection{Activity Start Time}

\begin{figure}[h]
  \centering
  \includegraphics[width=0.98\linewidth]{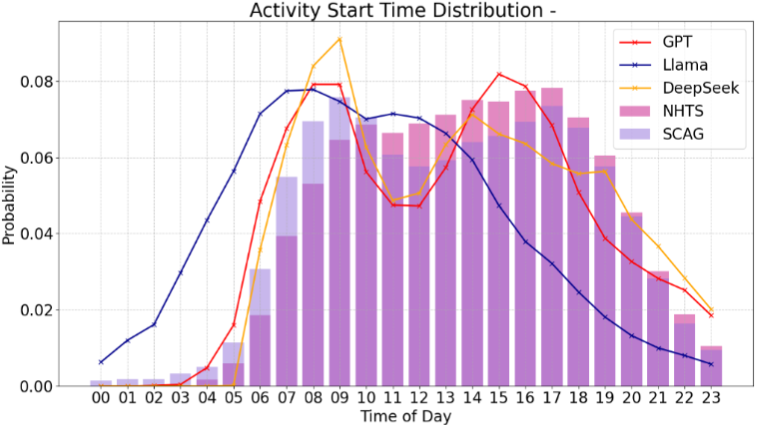}
  \caption{Start time distribution comparation}
  \label{fig:evaluation_overview_1}
\end{figure}

Fig.~\ref{fig:evaluation_overview_1} reveals our approach captures daily temporal patterns from both NHTS and SCAG datasets. The models reproduce characteristic peaks matching typical daily schedules. While all models capture general temporal trends, GPT-4o mini shows the closest match with reference patterns, especially for activity end times. DeepSeek excels at modeling early-day start times, highlighting how different architectures may capture specific temporal aspects of mobility.

\subsection{Activity pattern in different social group}

\begin{figure}[h]
  \centering
  \includegraphics[width=0.98\columnwidth]{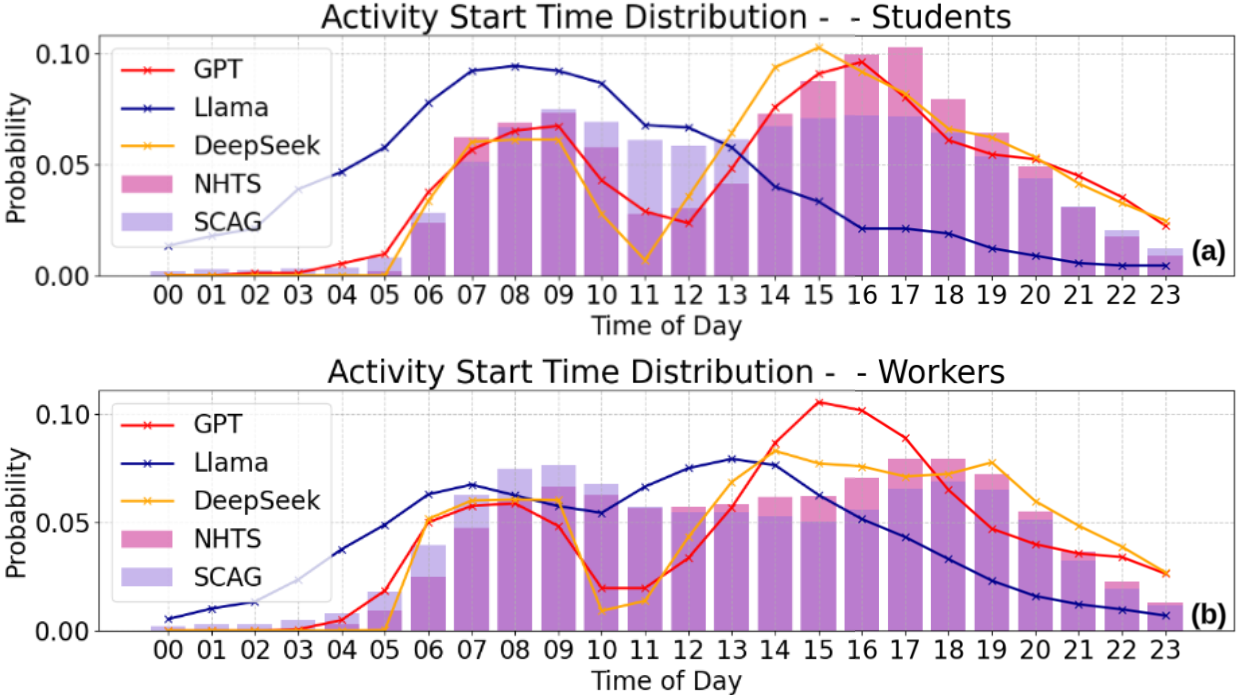}
  \caption{Activity start/end time distribution for students and workers}
  \label{fig:start_end_students_workers}
\end{figure}

Our detailed analysis of activity start times for specific social groups, namely students and workers as illustrated in Fig.~\ref{fig:start_end_students_workers}, demonstrates that our approach effectively captures distinct daily routines across different socio-demographic segments. The generated patterns successfully replicate characteristic peaks and time-of-day variations aligned with both NHTS and SCAG reference data. While GPT-4o-mini shows the best performance in modeling social group-specific mobility behaviors, our overall approach consistently represents realistic daily activity patterns for both students and workers. These results validate the effectiveness of our method in generating accurate human mobility patterns that reflect the temporal dynamics of different social groups.

\subsection{Activity Pattern by Activity Type}

\begin{figure}[h]
  \centering
  \includegraphics[width=0.98\columnwidth]{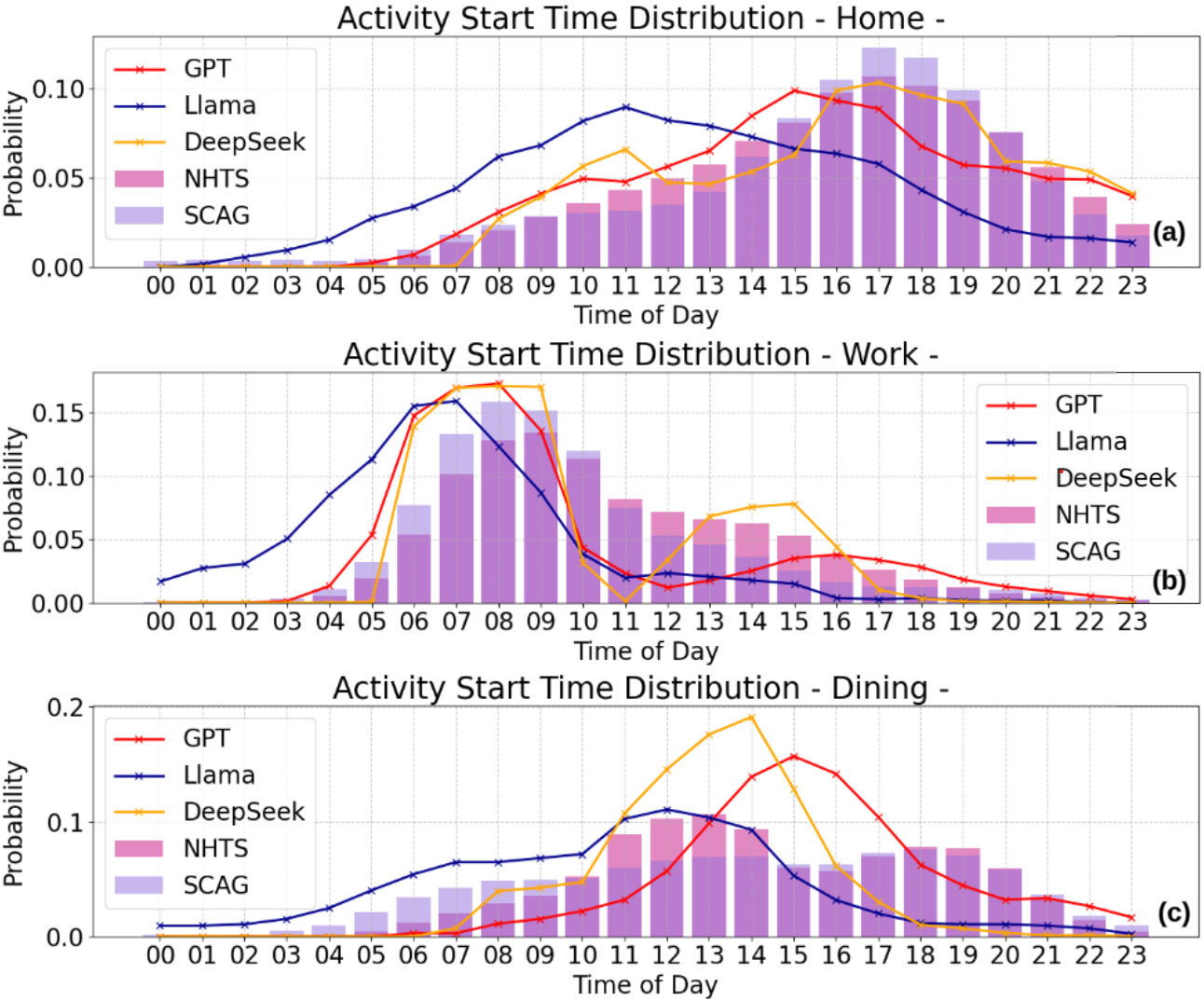}
  \caption{Activity start/end time distribution for home, work, and dining activities}
  \label{fig:start_activity_type}
\end{figure}

The analysis of activity start times categorized by activity types, as shown in Fig.~\ref{fig:start_activity_type}, highlights the robustness of our proposed approach in capturing nuanced temporal patterns. GPT-4o-mini demonstrates superior alignment with the reference datasets across all analyzed activity types, particularly in accurately reproducing the peak timing patterns for home, work, and dining activities. DeepSeek exhibits notable precision in capturing morning peaks for work-related activities, while Llama, though broadly accurate, tends to show more variability in capturing home and dining patterns. These findings emphasize the efficacy of our retrieval-augmented approach and underline GPT-4's particularly strong performance in modeling activity-specific timing trends.

\subsection{Household Coordination Activities}

\begin{figure}[h]
  \centering
  \includegraphics[width=0.98\columnwidth]{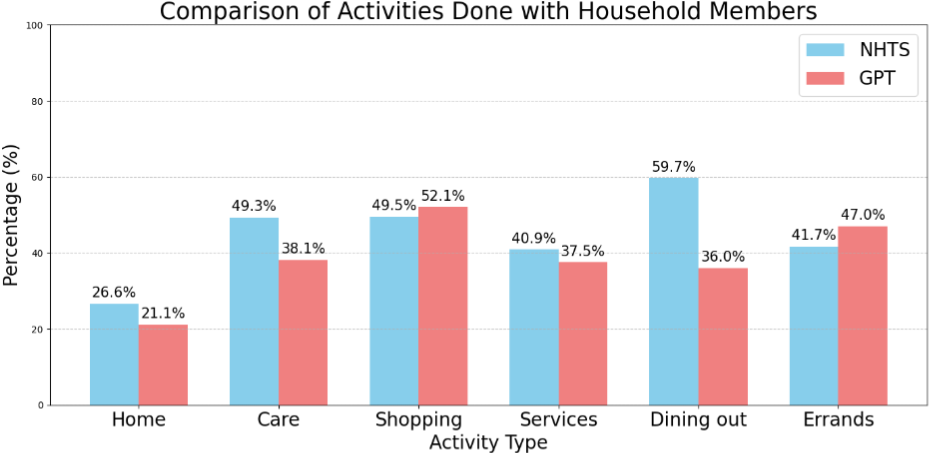}
  \caption{Household coordination activities participate rate compared with NHTS dataset}
  \label{fig:household_overview}
\end{figure}

Based on GPT-4o mini's superior performance in previous evaluations, we focused our household coordination analysis on results from this model. Our analysis demonstrates excellent overall alignment with the NHTS reference dataset as shown in Fig.~\ref{fig:household_overview}. The model successfully captures realistic patterns of joint household activities, particularly those where family members are most likely to participate together. We further analyzed specific relation pairs, namely head-spouse and head-child interactions, to evaluate the model's performance at a more granular level.

\begin{table}[h]
  \centering
  \caption{JSD values for household coordination activities participate rate compared with NHTS dataset}
  \label{table:household_coordination}
  \begin{tabular}{|>{\centering\arraybackslash}p{1.5cm}|>{\centering\arraybackslash}p{0.6cm}|>{\centering\arraybackslash}p{0.6cm}|>{\centering\arraybackslash}p{0.6cm}|>{\centering\arraybackslash}p{0.5cm}|>{\centering\arraybackslash}p{0.65cm}|>{\centering\arraybackslash}p{0.85cm}|}
    \hline
    \textbf{Relation} & \textbf{Home} & \textbf{Care} & \textbf{Shop} & \textbf{Serv.} & \textbf{Dining} & \textbf{Errands} \\
    \hline
    Head-Spouse & 0.0003 & 0.028 & 0.012 & 0.000 & 0.011 & 0.031 \\
    \hline
    Head-Child & 0.0003 & 0.049 & 0.005 & 0.005 & 0.057 & 0.011 \\
    \hline
  \end{tabular}
\end{table}

As presented in Table~\ref{table:household_coordination}, household-interactive activities such as home, shopping, and errands show particularly strong alignment for both relation types, with minimal divergence measured by low Jensen–Shannon divergence values. While care, dining out, and service-related activities—all common scenarios for joint household participation—exhibit slightly greater discrepancies, especially in head-child interactions, they remain within acceptable ranges. This detailed analysis by relation pair confirms that GPT-4o mini effectively reproduces household-coordinated behaviors across different family relationships, highlighting the strength of our retrieval-augmented approach in generating contextually consistent and realistic household interactions for the most common joint activities.

\section{Ablation Study and Hallucination Eliminating}

To evaluate the effectiveness of our retrieval-augmented feedback approach in generating consistent activity chains with household coordination and reducing hallucinations, we conducted an ablation study comparing the model's performance with and without this mechanism. 

\begin{table}[h]
  \centering
  \renewcommand{\arraystretch}{1.2}
  \caption{JSD values comparing activity chain statistics with and without retrieval-augmented feedback on statistical distributions on the length of generated activity chains.}
  \begin{tabular}{|>{\centering\arraybackslash}p{2.0cm}|>{\centering\arraybackslash}p{0.65cm}|>{\centering\arraybackslash}p{0.7cm}|>{\centering\arraybackslash}p{0.7cm}|>{\centering\arraybackslash}p{1.0cm}|>{\centering\arraybackslash}p{0.8cm}|}
  \hline
  Comparison & Type &  Time start & Time end & Duration & Length \\ \hline
  NHTS-GPT & 0.032 & 0.063 & 0.072 & 0.031 & 0.110 \\ \hline
  NHTS-GPT (Feedback)& \textbf{0.023} & \textbf{0.015} & \textbf{0.011} & \textbf{0.022} & \textbf{0.011} \\ \Xhline{3\arrayrulewidth}
  SCAG-GPT & 0.091 & 0.061 & 0.068 & 0.028 & 0.090 \\ \hline
  SCAG-GPT (Feedback)& \textbf{0.028} & \textbf{0.009} & \textbf{0.009} & \textbf{0.020} & \textbf{0.009} \\ \hline
  \end{tabular}
  \label{table:jsd_comparison_ab}
\end{table}

As shown in Table~\ref{table:jsd_comparison_ab}, our retrieval-augmented feedback mechanism provides statistical distributions on the length of generated activity chains. Using GPT-4o mini, the best-performing model, we observed substantial improvements across all metrics when testing on both NHTS and SCAG datasets. These results clearly demonstrate that our retrieval-augmented feedback mechanism effectively constrains the model to generate activity chains that more closely match real-world statistical distributions in terms of activity type, timing, duration, and length.

Moving beyond general statistical distributions, we also examine how our approach impacts household coordination activities. Table~\ref{table:ablation_household} demonstrates the critical role of our retrieval-augmented feedback mechanism in eliminating hallucinations and maintaining household activity consistency. With the feedback loop, 94.1\% of activities claimed to be performed with household members match corresponding activities by those members, compared to only 29.4\% without this mechanism.

\begin{table}[h]
  \centering
  \caption{Impact of Retrieval-Augmented Feedback on Household Activity Consistency}
  \label{table:ablation_household}
  \begin{tabular}{|>{\centering\arraybackslash}p{3.5cm}|c|c|}
    \hline
    \textbf{Approach} & \textbf{Consistent} & \textbf{Inconsistent} \\
    \hline
    With RAG Feedback & 1,011 (94.1\%) & 63 (5.9\%) \\
    \hline
    Without RAG Feedback & 509 (29.4\%) & 1,223 (70.6\%) \\
    \hline
  \end{tabular}
\end{table}

This difference highlights how our approach addresses a key challenge in activity chain generation: ensuring logical consistency across interdependent agents. The feedback mechanism reduces hallucinations while producing more realistic household dynamics by ensuring joint activities are actually shared among household members. 

Our hallucination elimination strategy creates a cross-verification system where household members' activities are checked against others' reported schedules, reducing hallucinated joint activities from 70.6\% to 5.9\%. This demonstrates our approach not only generates more accurate activity patterns but also effectively mitigates a primary concern in using LLMs for simulation tasks requiring logical consistency across multiple agents.

\section{Conclusion and Future Work}

This study presents a novel LLM-based approach for generating human mobility patterns using minimal socio-demographic data, leveraging GPT-4o mini and Llama2-70b with NHTS and SCAG datasets to accurately simulate daily activities. Our framework demonstrates strong alignment with real-world patterns through low JSD values, with GPT-4o mini excelling in activity type and duration modeling while our retrieval-augmented feedback mechanism reduced hallucinations in household coordination from 70.6\% to 5.9\%. While offering substantial benefits for urban planning through reduced data requirements and cross-contextual adaptability, limitations on generation efficiency and coverage still remain, future work will extend to multi-day forecasts with richer datasets, specialized fine-tuning techniques, and the development of hybrid models that combine learning-based methods with LLMs to enable low-cost, large-scale simulations, potentially establishing new benchmarks in mobility modeling for diverse urban environments while maintaining computational feasibility for widespread deployment.

\section*{Acknowledgment}
This work was supported by the Intelligence Advanced Research Projects Activity (IARPA) via the Department of Interior/Interior Business Center (DOI/IBC) contract number 140D0423C0033. The U.S. Government is authorized to reproduce and distribute reprints for Governmental purposes notwithstanding any copyright annotation thereon. Disclaimer: The views and conclusions contained herein are those of the authors and should not be interpreted as necessarily representing the official policies or endorsements, either expressed or implied, of IARPA, DOI/IBC, or the U.S. Government.

% \clearpage
\bibliographystyle{IEEEtran}
\bibliography{itsc_2025_activity_chain_generation}

\begin{thebibliography}{10}
\providecommand{\url}[1]{#1}
\csname url@samestyle\endcsname
\providecommand{\newblock}{\relax}
\providecommand{\bibinfo}[2]{#2}
\providecommand{\BIBentrySTDinterwordspacing}{\spaceskip=0pt\relax}
\providecommand{\BIBentryALTinterwordstretchfactor}{4}
\providecommand{\BIBentryALTinterwordspacing}{\spaceskip=\fontdimen2\font plus
\BIBentryALTinterwordstretchfactor\fontdimen3\font minus \fontdimen4\font\relax}
\providecommand{\BIBforeignlanguage}[2]{{%
\expandafter\ifx\csname l@#1\endcsname\relax
\typeout{** WARNING: IEEEtran.bst: No hyphenation pattern has been}%
\typeout{** loaded for the language `#1'. Using the pattern for}%
\typeout{** the default language instead.}%
\else
\language=\csname l@#1\endcsname
\fi
#2}}
\providecommand{\BIBdecl}{\relax}
\BIBdecl

\bibitem{barbosa2018human}
H.~Barbosa, M.~Barthelemy, G.~Ghoshal, C.~R. James, M.~Lenormand, T.~Louail, R.~Menezes, J.~J. Ramasco, F.~Simini, and M.~Tomasini, ``Human mobility: Models and applications,'' \emph{Physics Reports}, vol. 734, pp. 1--74, 2018.

\bibitem{li2021impact}
Y.~Li, M.~Li, M.~Rice, H.~Zhang, D.~Sha, M.~Li, Y.~Su, and C.~Yang, ``The impact of policy measures on human mobility, covid-19 cases, and mortality in the us: a spatiotemporal perspective,'' \emph{International Journal of Environmental Research and Public Health}, vol.~18, no.~3, p. 996, 2021.

\bibitem{vanhaverbeke2011agent}
L.~Vanhaverbeke and C.~Macharis, ``An agent-based model of consumer mobility in a retail environment,'' \emph{Procedia-Social and Behavioral Sciences}, vol.~20, pp. 186--196, 2011.

\bibitem{rasouli2014activity}
S.~Rasouli and H.~Timmermans, ``Activity-based models of travel demand: promises, progress and prospects,'' \emph{International Journal of Urban Sciences}, vol.~18, no.~1, pp. 31--60, 2014.

\bibitem{bhat1999activity}
C.~R. Bhat and F.~S. Koppelman, ``Activity-based modeling of travel demand,'' in \emph{Handbook of transportation Science}.\hskip 1em plus 0.5em minus 0.4em\relax Springer, 1999, pp. 35--61.

\bibitem{goulias2011simulator}
K.~G. Goulias, C.~R. Bhat, R.~M. Pendyala, Y.~Chen, R.~Paleti, K.~C. Konduri, G.~Huang, and H.-H. Hu, ``Simulator of activities, greenhouse emissions, networks, and travel (simagent) in southern california: Design, implementation, preliminary findings, and integration plans,'' in \emph{2011 IEEE Forum on Integrated and Sustainable Transportation Systems}.\hskip 1em plus 0.5em minus 0.4em\relax IEEE, 2011, pp. 164--169.

\bibitem{mcfadden1974measurement}
D.~McFadden, ``The measurement of urban travel demand,'' \emph{Journal of public economics}, vol.~3, no.~4, pp. 303--328, 1974.

\bibitem{heffer2021impact}
H.~Heffer-Flaata, A.~Voltes-Dorta, and P.~Suau-Sanchez, ``The impact of accommodation taxes on outbound travel demand from the united kingdom to european destinations,'' \emph{Journal of Travel Research}, vol.~60, no.~4, pp. 749--760, 2021.

\bibitem{sila2016analysis}
K.~Si{\l}a-Nowicka, J.~Vandrol, T.~Oshan, J.~A. Long, U.~Dem{\v{s}}ar, and A.~S. Fotheringham, ``Analysis of human mobility patterns from gps trajectories and contextual information,'' \emph{International Journal of Geographical Information Science}, vol.~30, no.~5, pp. 881--906, 2016.

\bibitem{huang2018modeling}
Z.~Huang, X.~Ling, P.~Wang, F.~Zhang, Y.~Mao, T.~Lin, and F.-Y. Wang, ``Modeling real-time human mobility based on mobile phone and transportation data fusion,'' \emph{Transportation research part C: emerging technologies}, vol.~96, pp. 251--269, 2018.

\bibitem{tang2015uncovering}
J.~Tang, F.~Liu, Y.~Wang, and H.~Wang, ``Uncovering urban human mobility from large scale taxi gps data,'' \emph{Physica A: Statistical Mechanics and its Applications}, vol. 438, pp. 140--153, 2015.

\bibitem{liao2024deep}
X.~Liao, Q.~Jiang, B.~Y. He, Y.~Liu, C.~Kuai, and J.~Ma, ``Deep activity model: A generative approach for human mobility pattern synthesis,'' \emph{arXiv preprint arXiv:2405.17468}, 2024.

\bibitem{chakraborty2017interpretability}
S.~Chakraborty, R.~Tomsett, R.~Raghavendra, D.~Harborne, M.~Alzantot, F.~Cerutti, M.~Srivastava, A.~Preece, S.~Julier, R.~M. Rao \emph{et~al.}, ``Interpretability of deep learning models: A survey of results,'' in \emph{2017 IEEE smartworld, ubiquitous intelligence \& computing, advanced \& trusted computed, scalable computing \& communications, cloud \& big data computing, Internet of people and smart city innovation (smartworld/SCALCOM/UIC/ATC/CBDcom/IOP/SCI)}.\hskip 1em plus 0.5em minus 0.4em\relax IEEE, 2017, pp. 1--6.

\bibitem{pellungrini2017fast}
R.~Pellungrini, L.~Pappalardo, F.~Pratesi, and A.~Monreale, ``Fast estimation of privacy risk in human mobility data,'' in \emph{Computer Safety, Reliability, and Security: SAFECOMP 2017 Workshops, ASSURE, DECSoS, SASSUR, TELERISE, and TIPS, Trento, Italy, September 12, 2017, Proceedings 36}.\hskip 1em plus 0.5em minus 0.4em\relax Springer, 2017, pp. 415--426.

\bibitem{ma2024mobility}
H.~Ma, Y.~Liu, Q.~Jiang, B.~Y. He, X.~Liao, and J.~Ma, ``Mobility ai agents and networks,'' \emph{IEEE Transactions on Intelligent Vehicles}, 2024.

\bibitem{liu2024semantic}
Y.~Liu, C.~Kuai, H.~Ma, X.~Liao, B.~Y. He, and J.~Ma, ``Semantic trajectory data mining with llm-informed poi classification,'' \emph{arXiv preprint arXiv:2405.11715}, 2024.

\bibitem{achiam2023gpt}
J.~Achiam, S.~Adler, S.~Agarwal, L.~Ahmad, I.~Akkaya, F.~L. Aleman, D.~Almeida, J.~Altenschmidt, S.~Altman, S.~Anadkat \emph{et~al.}, ``Gpt-4 technical report,'' \emph{arXiv preprint arXiv:2303.08774}, 2023.

\bibitem{NEURIPS2020_1457c0d6}
T.~Brown, B.~Mann, N.~Ryder, M.~Subbiah, J.~D. Kaplan, P.~Dhariwal, A.~Neelakantan, P.~Shyam, G.~Sastry, A.~Askell, S.~Agarwal, A.~Herbert-Voss, G.~Krueger, T.~Henighan, R.~Child, A.~Ramesh, D.~Ziegler, J.~Wu, C.~Winter, C.~Hesse, M.~Chen, E.~Sigler, M.~Litwin, S.~Gray, B.~Chess, J.~Clark, C.~Berner, S.~McCandlish, A.~Radford, I.~Sutskever, and D.~Amodei, ``Language models are few-shot learners,'' in \emph{Advances in Neural Information Processing Systems}, H.~Larochelle, M.~Ranzato, R.~Hadsell, M.~Balcan, and H.~Lin, Eds., vol.~33.\hskip 1em plus 0.5em minus 0.4em\relax Curran Associates, Inc., 2020, pp. 1877--1901.

\bibitem{NIPS2017_3f5ee243}
A.~Vaswani, N.~Shazeer, N.~Parmar, J.~Uszkoreit, L.~Jones, A.~N. Gomez, L.~u. Kaiser, and I.~Polosukhin, ``Attention is all you need,'' in \emph{Advances in Neural Information Processing Systems}, I.~Guyon, U.~V. Luxburg, S.~Bengio, H.~Wallach, R.~Fergus, S.~Vishwanathan, and R.~Garnett, Eds., vol.~30.\hskip 1em plus 0.5em minus 0.4em\relax Curran Associates, Inc., 2017.

\bibitem{stouffer1940intervening}
S.~A. Stouffer, ``Intervening opportunities: a theory relating mobility and distance,'' \emph{American sociological review}, vol.~5, no.~6, pp. 845--867, 1940.

\bibitem{zipf1941national}
G.~K. Zipf, ``National unity and disunity; the nation as a bio-social organism.'' 1941.

\bibitem{bhat2012household}
C.~R. Bhat, K.~G. Goulias, R.~M. Pendyala, R.~Paleti, R.~Sidharthan, L.~Schmitt, and H.-h. Hu, ``A household-level activity pattern generation model for the simulator of activities, greenhouse emissions, networks, and travel (simagent) system in southern california,'' in \emph{91st Annual Meeting of the Transportation Research Board, Washington, DC}, 2012.

\bibitem{bhat2013household}
C.~R. Bhat, K.~G. Goulias, R.~M. Pendyala, R.~Paleti, R.~Sidharthan, L.~Schmitt, and H.-H. Hu, ``A household-level activity pattern generation model with an application for southern california,'' \emph{Transportation}, vol.~40, pp. 1063--1086, 2013.

\bibitem{kim2018method}
D.~Y. Kim and H.~Y. Song, ``Method of predicting human mobility patterns using deep learning,'' \emph{Neurocomputing}, vol. 280, pp. 56--64, 2018.

\bibitem{kong2022exploring}
X.~Kong, K.~Wang, M.~Hou, F.~Xia, G.~Karmakar, and J.~Li, ``Exploring human mobility for multi-pattern passenger prediction: A graph learning framework,'' \emph{IEEE Transactions on Intelligent Transportation Systems}, vol.~23, no.~9, pp. 16\,148--16\,160, 2022.

\bibitem{hoteit2014estimating}
S.~Hoteit, S.~Secci, S.~Sobolevsky, C.~Ratti, and G.~Pujolle, ``Estimating human trajectories and hotspots through mobile phone data,'' \emph{Computer Networks}, vol.~64, pp. 296--307, 2014.

\bibitem{yin2011diversified}
Z.~Yin, L.~Cao, J.~Han, J.~Luo, and T.~Huang, ``Diversified trajectory pattern ranking in geo-tagged social media,'' in \emph{Proceedings of the 2011 SIAM international conference on data mining}.\hskip 1em plus 0.5em minus 0.4em\relax SIAM, 2011, pp. 980--991.

\bibitem{srinivasan2005modeling}
S.~Srinivasan and C.~R. Bhat, ``Modeling household interactions in daily in-home and out-of-home maintenance activity participation,'' \emph{Transportation}, vol.~32, pp. 523--544, 2005.

\bibitem{rezvany2023simulating}
N.~Rezvany, M.~Bierlaire, and T.~Hillel, ``Simulating intra-household interactions for in-and out-of-home activity scheduling,'' \emph{Transportation Research Part C: Emerging Technologies}, vol. 157, p. 104362, 2023.

\bibitem{puig2018virtualhome}
X.~Puig, K.~Ra, M.~Boben, J.~Li, T.~Wang, S.~Fidler, and A.~Torralba, ``Virtualhome: Simulating household activities via programs,'' in \emph{Proceedings of the IEEE conference on computer vision and pattern recognition}, 2018, pp. 8494--8502.

\bibitem{albouys2019smach}
J.~Albouys, N.~Sabouret, Y.~Haradji, M.~Schumann, and C.~Inard, ``Smach: Multi-agent simulation of human activity in the household,'' in \emph{Advances in Practical Applications of Survivable Agents and Multi-Agent Systems: The PAAMS Collection: 17th International Conference, PAAMS 2019, {\'A}vila, Spain, June 26--28, 2019, Proceedings 17}.\hskip 1em plus 0.5em minus 0.4em\relax Springer, 2019, pp. 227--231.

\bibitem{li2024personal}
Y.~Li, H.~Wen, W.~Wang, X.~Li, Y.~Yuan, G.~Liu, J.~Liu, W.~Xu, X.~Wang, Y.~Sun \emph{et~al.}, ``Personal llm agents: Insights and survey about the capability, efficiency and security,'' \emph{arXiv preprint arXiv:2401.05459}, 2024.

\bibitem{mao2023gpt}
J.~Mao, Y.~Qian, H.~Zhao, and Y.~Wang, ``Gpt-driver: Learning to drive with gpt,'' \emph{arXiv preprint arXiv:2310.01415}, 2023.

\bibitem{gao2023retrieval}
Y.~Gao, Y.~Xiong, X.~Gao, K.~Jia, J.~Pan, Y.~Bi, Y.~Dai, J.~Sun, H.~Wang, and H.~Wang, ``Retrieval-augmented generation for large language models: A survey,'' \emph{arXiv preprint arXiv:2312.10997}, vol.~2, 2023.

\bibitem{ni2025towards}
B.~Ni, Z.~Liu, L.~Wang, Y.~Lei, Y.~Zhao, X.~Cheng, Q.~Zeng, L.~Dong, Y.~Xia, K.~Kenthapadi \emph{et~al.}, ``Towards trustworthy retrieval augmented generation for large language models: A survey,'' \emph{arXiv preprint arXiv:2502.06872}, 2025.

\bibitem{yu2023refeed}
W.~Yu, Z.~Zhang, Z.~Liang, M.~Jiang, and A.~Sabharwal, ``Improving language models via plug-and-play retrieval feedback,'' \emph{arXiv preprint arXiv:2305.14002}, 2023.

\bibitem{peng2023check}
B.~Peng, M.~Galley, P.~He, H.~Cheng, Y.~Xie, Y.~Hu, Q.~Huang, L.~Liden, Z.~Yu, W.~Chen, and J.~Gao, ``Check your facts and try again: Improving large language models with external knowledge and automated feedback,'' \emph{arXiv preprint arXiv:2302.12813}, 2023.

\bibitem{fish2015transportation}
J.~Fish, J.~Gonder, V.~Garikapati, J.~Cappellucci, B.~Borlaug, J.~Holden, and L.~Boyce, ``Transportation secure data center,'' DOE Open Energy Data Initiative (OEDI); National Renewable Energy Laboratory~…, Tech. Rep., 2015.

\bibitem{NHTS2009}
{U.S. Department of Transportation, Federal Highway Administration}, ``2009 national household travel survey,'' \url{https://nhts.ornl.gov}, 2009.

\bibitem{he2022connected}
B.~Y. He, Q.~Jiang, and J.~Ma, ``Connected automated vehicle impacts in southern california part-i: Travel behavior and demand analysis,'' \emph{Transportation research part D: transport and environment}, vol. 109, p. 103329, 2022.

\bibitem{jiang2022connected}
Q.~Jiang, B.~Y. He, and J.~Ma, ``Connected automated vehicle impacts in southern california part-ii: Vmt, emissions, and equity,'' \emph{Transportation research part D: transport and environment}, vol. 109, p. 103381, 2022.

\bibitem{menendez1997jensen}
M.~L. Men{\'e}ndez, J.~Pardo, L.~Pardo, and M.~Pardo, ``The jensen-shannon divergence,'' \emph{Journal of the Franklin Institute}, vol. 334, no.~2, pp. 307--318, 1997.

\end{thebibliography}

\end{document}